\pdfoutput=1
\documentclass{article}
\usepackage[numbers]{natbib}
\usepackage[final,nonatbib]{nips_2018}
\usepackage{hyperref}
\usepackage{url}
\usepackage{cleveref}
\usepackage{subcaption}
\newcommand{\rulesep}{\unskip\ \vrule\ }
\usepackage{algorithm}
\usepackage{algorithmic}
\usepackage{times,amsmath,amssymb,amsthm,graphicx,url,xspace,wrapfig,mathtools,caption}

\newenvironment{itemize*}%
  {\begin{itemize}%
    \setlength{\itemsep}{0pt}%
    \setlength{\parskip}{0pt}}%
  {\end{itemize}}
\newenvironment{enumerate*}%
  {\begin{enumerate}%
    \setlength{\itemsep}{0pt}%
    \setlength{\parskip}{0pt}}%
  {\end{enumerate}}

\newtheorem{definition}{Definition}

\newcommand{\x}{{\boldsymbol{x}}}

\newcommand{\RR}{\mathbb{R}}

\newcommand{\Fcal}{{\mathcal{F}}}
\newcommand{\Xcal}{{\mathcal{X}}}

\newcommand{\Hcal}{{\mathcal{H}}}

\newcommand{\bx}{{\boldsymbol{x}}}
\newcommand{\bc}{{\boldsymbol{c}}}

\newcommand{\bv}{{\boldsymbol{v}}}

\usepackage{color}

\ifdefined\QED
\else
\newcommand{\QED}{\hfill \ensuremath{\Box}}
\fi

\title{Unsupervised Detection and Explanation of Latent-class Contextual Anomalies}
\author{
Nico G\"{o}rnitz \\
Machine Learning Group\\
TU Berlin \\
\texttt{nico.goernitz@tu-berlin.de} \\
\And
Luiz Alberto Lima\\
Petrobras, 20031-912 Rio de Janeiro, Brazil\\
\texttt{lual@petrobras.com.br} \\
\And
Shinichi Nakajima \\
Machine Learning Group\\
TU Berlin \\
\texttt{shinichi.nakajima@tu-berlin.de} \\
\And
Marius Kloft$^*$\\
Department of Computer Science\\
HU Berlin\\
\texttt{kloft@hu-berlin.de} 
}

\author{
Jacob Kauffmann$^{1}$, Gr\'{e}goire Montavon$^1$,
Luiz Alberto Lima$^{2}$\\
{\bf Shinichi Nakajima$^{1}$, Klaus-Robert M{\"u}ller$^{1,3,4,5}$ and Nico G\"{o}rnitz$^{1}$} \\
$^1$TU Berlin, $^2$Petrobras, \\
$^3$AIP, RIKEN, $^4$Korea University, $^5$MPI for Informatics \\
\texttt{\{j.kauffmann,gregoire.montavon,nakajima,klaus-robert.mueller, }\\
 \texttt{\qquad nico.goernitz\}@tu-berlin.de }\\
\texttt{lual@petrobras.com.br}
}

\begin{document}
\maketitle

%--------------------------------------------
\begin{abstract}
Detecting and explaining anomalies is a challenging effort. This holds especially true when data exhibits strong dependencies and single measurements need to be assessed and analyzed in their respective context. In this work, we consider scenarios where measurements are non-i.i.d, i.e.\ where samples are dependent on corresponding discrete latent variables which are connected through some given dependency structure, the contextual information. 
Our contribution is twofold: (i) Building atop of support vector data description (SVDD), we derive a method able to cope with latent-class dependency structure that can still be optimized efficiently. We further show that our approach neatly generalizes vanilla SVDD as well as $k$-means and conditional random fields (CRF) and provide a corresponding probabilistic interpretation. (ii) In unsupervised scenarios where it is not possible to quantify the accuracy of an anomaly detector, having an human-interpretable solution is the key to success. Based on deep Taylor decomposition and a reformulation of our trained anomaly detector as a neural network, we are able to backpropagate predictions to pixel-domain and thus identify features and regions of high relevance.
We demonstrate the usefulness of our novel approach on toy data with known spatio-temporal structure and successfully validate on synthetic as well as real world off-shore data from the oil industry.
\end{abstract}
%--------------------------------------------
%
%
%--------------------------------------------
\section{Introduction}\label{sec:intro}
%--------------------------------------------
Addressing complex outliers is an important challenge for machine learning and statistics. Outliers need to be detected and removed as they typically prevent learning systems to generalize \cite{Huber1981}. On the other side an outlier itself can hold important information and thus be the central target of interest in data analysis \cite{Harmeling2006,Chandola2009,AggarwalOutlier2013}. Especially complex outliers are hard to detect since the underlying data may be structured \cite{GoeBraKlo15,Song2013,HoeNakBauMueGoe17} and/or has unknown latent variables \cite{Porbadnigk2015} and it is precisely this new and challenging scenario where this work will contribute with a novel mathematical model.

\begin{wrapfigure}{r}{0.4\textwidth}
  \begin{center}
	\includegraphics[width=0.37\textwidth]{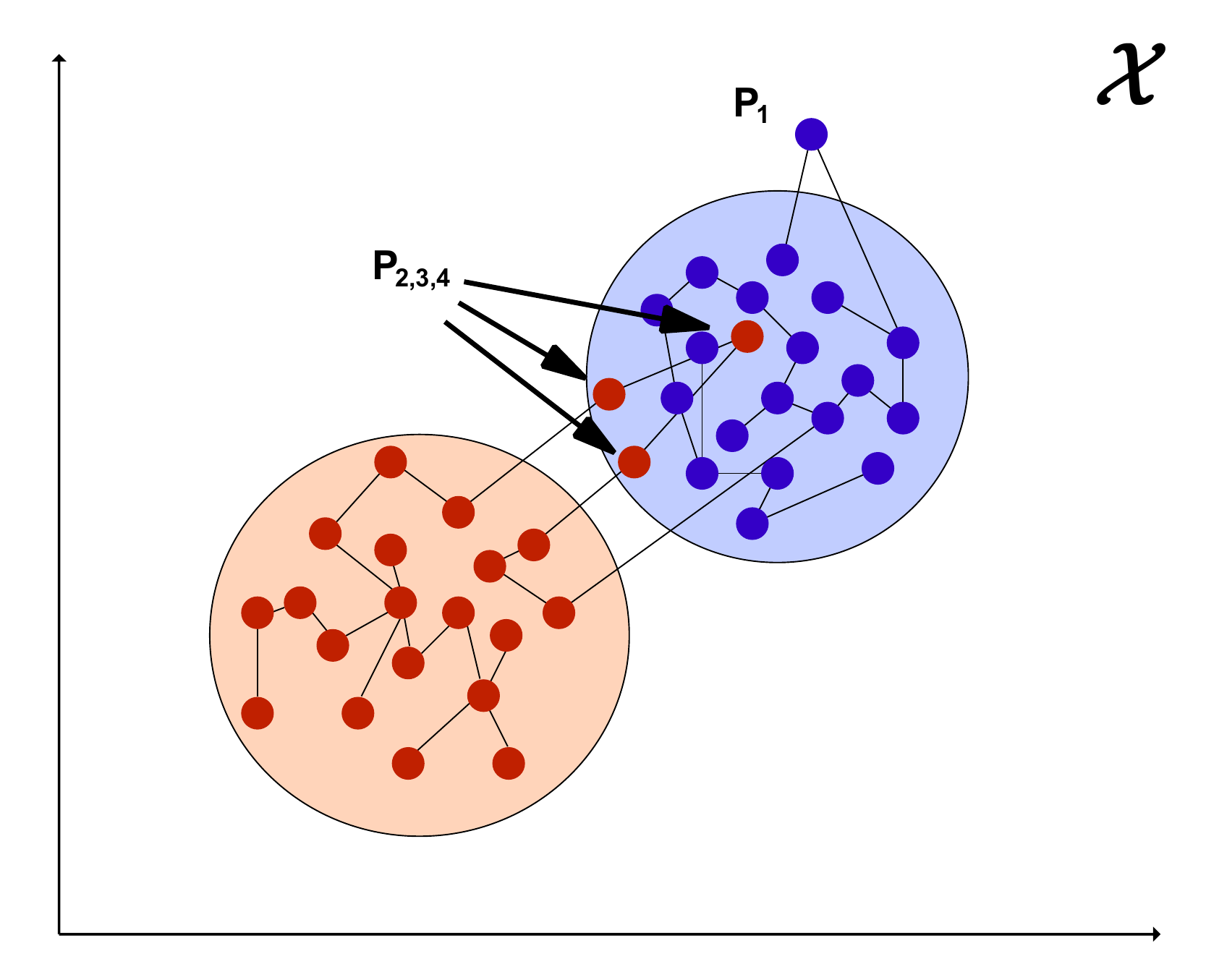}
\caption{Our proposed model LCCAD. }
\label{fig:lccad-model}
  \end{center}
 \vspace{-1cm}
\end{wrapfigure}

Specifically we propose a CRF-based \cite{LafMccPer01} one-class classifier \cite{SchPlaShaSmoWil01,TaxDui04} that incorporates latent structure. Figure~\ref{fig:lccad-model} illustrates the idea  of our proposed method that we would like to call latent-class contextual anomaly detector (LCCAD). Here, we assume an unsupervised scenario where the data is generated from two unknown latent processes (red and blue dots in the figure). We would like to learn two hyperspheres (normal data) containing most of the data points corresponding to the respective latent variables (color) while taking the given (complex) dependency structure (black lines) into account. This should allow us to detect anomalies even if they hide in high density regions (cf. points $P_{2,3,4}$). This is in stark contrast to traditional anomaly detectors such as vanilla SVDD \cite{TaxDui04}, one-class SVM \cite{SchPlaShaSmoWil01}, isolation forest \cite{LiuTinZho08,LiuTinZho12}, and LODA \cite{Pevny2016} or structured approaches such as \cite{GoeBraKlo15,Song2013} where {\em no} unknown latent variable is present beyond existing structures.

While in toy scenarios we can measure and assess the quality of our model’s performance, analyzing complex real-world data requires more: {\em Explanation} needs to be provided to a user why a certain specific data point is considered anomalous e.g.\ by answering which input features of the data point make the model decide ``anomalous''.

We provide toy simulations demonstrating that our model, unlike existing ones can detect reliably for this scenario. To clearly demonstrate that the described complex structured outlier detection scenario is a practically existing important real-world problem, we study Geophysics data from oil exploration; also here we observe that our LCCAD algorithm compares very favorably to potential competing approaches. In addition we show the usefulness of our novel explanation method for LCCAD anomaly detection of geophysical facies data and illustrate possible scientific insights that can be obtained.
In the following, we will introduce our novel proposed LCCAD method in Section~\ref{sec:method} and the corresponding explanation methodology in Section~\ref{sec:explanation}. Section~\ref{sec:experiments} presents empirical results on toy data as well as on synthetic and real-world geophysics data. We conclude with Section~\ref{sec:conclusion}. 

%--------------------------------------
\section{Detecting Latent-class Contextual Anomalies}\label{sec:method}
%--------------------------------------

The mathematical set-up of our unsupervised learning probem is the following: We are given an unlabeled dataset $X := \{\bx_1, \ldots, \bx_n\}$ with $\bx \in \Xcal$,
a feature map $\phi: \Xcal \rightarrow \Fcal$ mapping each data point into a possibly high dimensional feature space, and a general loss function $\ell: \RR \rightarrow \RR$. Further, each entry $\bx_i$ is assigned a corresponding discrete latent class variable $h_i \in \{1,\ldots,K\}$ with $H:=\{h_1,\ldots,h_n\}$. The task is to find the center 
$\bc_k \in \Fcal$ and the radius $\sqrt{t_k}$ of a hypersphere for each context $k \in \{1,\ldots,K\}$ such that the bulk of the corresponding data points belonging to this context $X_k := \{\bx_i | h_i=k, \, i=1,\ldots,n\}$ is contained and a fraction $\nu$ lying outside or on the border. The set of centers and radii is referred to as $C:=\{\bc_k|k=1,\ldots,K\}$ and $T:=\{t_k|k=1,\ldots,K\}$ correspondingly.
Dependency structure between latent variables $h_i$ is induced using a joint feature map $\psi: \Xcal^n \times \{1,\ldots,K\}^n \rightarrow \Hcal$. 

Subsequently, we will derive our novel latent-class contextual anomaly detector (LCCAD), based on the support vector data description (SVDD) \cite{TaxDui04} and give its probabilistic interpretation. Additionally we will show how to induce structural dependencies among latent variables using Markov random fields (MRF) as an example. Further, we will discuss in detail how to efficiently solve the resulting optimization problems and consider special cases and relations to existing methods.

%--------------------------------------
\subsection{From SVDD to Latent-class Contextual Anomaly Detector}\label{sec:method}
%--------------------------------------
Support vector data description (SVDD) \cite{TaxDui04,Banerjee2006,GoeKloRieBre13} is a well-known efficient anomaly detector, assuming, however, i.i.d. data without any latent variable dependencies or other complex dependency structure. The formulation of unconstrained SVDD in \cite[Def.\ 3]{GoeLimMueKloNak2017} will now be extended to allow arbitrary loss functions to obtain ultimately our latent-class contextual anomaly detector (LCCAD)
\begin{align*}
\min_{\bc \in \Fcal, t\geq 0} f_\nu(\bc, t, X) = t + \frac{1}{n \nu} \sum_{i=1}^n \ell (\|\bc - \phi(\bx_i)\|^2 - t).
\end{align*}
Here, setting $\ell(x) = \max(0, x)$ recovers the Hinge loss formulation of the vanilla unconstrained SVDD.
Moreover, we extend this model in the spirit of ClusterSVDD in \cite[Def.\ 4]{GoeLimMueKloNak2017} by introducing latent variables $h_i \in \{1,\ldots,K\}$ for each example. Basically, the class assignment of each entry is selecting the corresponding SVDD. The resulting optimization neatly splits into a sum of $K$ independent SVDDs \emph{once the set of latent variables H is fixed}:
\begin{align*}
\min_{C \in \Fcal^K, T\geq 0, H \in \{1,\ldots,K\}^n} \sum_{k=1}^K t_k + \frac{1}{n_k \nu} \sum_{\{i|h_i=k\}} \ell (\|\bc_{h_i} - \phi(\bx_i)\|^2 - t_{h_i}) = \sum_{k=1}^K f_\nu(\bc_k, t_k, X_k)
\end{align*}
Finally, we respect the dependency structure between the latent variables by using a convex combination of the above optimization problem with a log-linear model employing joint feature maps. The specifics of the structure are hereby encoded in the joint feature map $\psi$. We give an example on the Markov random field in Section~\ref{sec:structure}.   
\begin{definition}[Latent-class Contextual Anomaly Detector, LCCAD]\label{def:lccad} Given pre-defined parameters $0 \leq \theta \leq 1$, the fraction of outliers $0 < \nu \leq 1$, and the regularizer trade-off $\gamma \geq 0$, the primal non-convex optimization problem of our proposed latent-class contextual anomaly detector (LCCAD) is given by:
\begin{align}\label{eq:lccad}
\min_{C \in \Fcal^K, T\geq 0, H \in \{1,\ldots,K\}^n} &\theta \left( \sum_{k=1}^K t_k + \frac{1}{n_k \nu} \sum_{\{i|h_i=k\}} \ell (\|\bc_{h_i} - \phi(\bx_i)\|^2 - t_{h_i}) \right) \\ 
& \; + (1-\theta) \left( \frac{\gamma}{2} \|\bv\|^2
 -   \langle \bv,\psi(X, H)\rangle + \log Z(X,\bv) \right), \nonumber
\end{align}
where $Z(X,\bv) = \sum_{\hat{H} \in \{1,\ldots,K\}^n } \exp (\langle \bv, \psi(X, \hat{H}) \rangle) $ denotes the partition function.
\end{definition}

%--------------------------------------
\subsection{A Probabilistic Interpretation}\label{sec:probabilistic}
%--------------------------------------
For the specific setting of LCCAD as given in Def.~\ref{def:lccad}, namely for $\nu = 1$ and $\ell$ being the Hinge-loss, we can derive 
a simple probabilistic interpretation in terms of mixture of Gaussians (with additional dependency structure). 
Setting $\ell(x) = \max(0, x)$ and $\nu = 1$ will result in $T_k = 0 \; \forall \,k$ (cf. Lemma 3 in \cite{GoeLimMueKloNak2017}) and the optimal parameters for the SVDD part can be solved analytically by $\bc_k = \frac{1}{n_k} \sum_{\{i|h_i=k\}} \phi(\bx_i)$ (cf. Theorem 2 in \cite{GoeLimMueKloNak2017}). 
We arrive at
\begin{align}\label{eq:simplified-model}
\min_{\bv, \bc \in \Fcal, H \in \{1,\ldots,K\}^n} &\theta \left( \sum_{k=1}^K \sum_{\{i|h_i=k\}} \|\bc_{h_i} - \phi(\bx_i)\|^2) \right)  \\
 &\; + (1-\theta) \left( \frac{\gamma}{2} \|\bv\|^2
 -   \langle \bv,\psi(X, H)\rangle + \log Z(X,\bv) \right) . \nonumber
\end{align}
The above optimization problem is a combination of a mixture model (much like $k$-means) and a conditional random field where the corresponding probabilistic model of the latter can be written as
\begin{align*}
% MODEL
p(X,H|\bc, \bv) &\sim \prod_i \prod_k \mathcal{N}(\bx_i | \bc_k, \sigma^2\mathbf{I})^{\mathbf{1}[k=h_i]} \cdot \frac{\exp(\langle \bv, \psi(X,H) \rangle)}{Z(X,\bv)} 
% PRIOR 
\quad \text{and} \quad
 p(\bv) \sim \mathcal{N}(\mathbf{0}, \gamma^{-1} \mathbf{I}).
\end{align*}
We are interested in finding $\max_{C} p(X|C) \geq \max_{C,H,\bv} p(X,H,\bv|C) = \max_{C,H,\bv} p(X,H|C,\bv)p(\bv)$, hence $\min_{C,H,\bv} -\log p(X,H|C,\bv) - \log p(\bv)$ (cf. Eqn~\eqref{eq:simplified-model}).

%--------------------------------------
\subsection{Inducing Structural Dependencies}\label{sec:structure}
%--------------------------------------
Here, we give an example of how to induce structural dependencies in case of Markov random fields (MRF), more specifically, conditional random fields (CRF) with the only difference between both being the log partition function: $Z(\bv)$ for the MRF and $Z(X,\bv)$ for the CRF.
Given a undirected graph $G = (V, E)$ with binary edges $E$ and vertices $V$, where each vertex corresponds to a sample and the state space is $S=\{1,\ldots,K\}$, we employ the following joint feature map:
\begin{align*}
\psi(X, H) &= 
        \left(
          \begin{array}{l}
     ( \sum_{(e_1, e_2) \in E} \mathbf{1}[h_{e_1}=s_1 \wedge h_{e_2}=s_2] )_{(s_1,s_2) \in S} ,\\
     \hspace{1.4cm}( \sum_{v \in V} \mathbf{1}[ h_v=s ] \, \phi(\bx_v) )_{s \in S}            
          \end{array}
   \right)
\end{align*}
and hence, 
\begin{align*}
P(H|X,\bv) = \frac{\exp(\langle \bv, \psi(X,H) \rangle)}{Z(X,\bv)}  \propto
\prod_{(e_1,e_2) \in E} \underbrace{\exp(\bv^{trans}_{(h_{e_1},h_{e_2})}) }_{\Psi_{(e_1,e_2)}(h_{e_1}, h_{e_2}; \bv)}
\prod_{v \in V} \underbrace{\exp(\langle \bv^{em}_{h_v}, \phi(\bx_v) \rangle)}_{\Psi_v(h_v; \bx_v, \bv)} .
\end{align*}
Joint feature maps had been introduced in \cite{MccFrePer00,Collins2002} and are heavily used in structured output prediction, e.g. \cite{TsoJoaHofAlt05}. 

%--------------------------------------
\subsection{Efficient Optimization}
%--------------------------------------
The key for our model to be applicable in practice is an efficient optimization of the non-convex problem as stated in Def.~\ref{def:lccad}. A common scheme is to alternate between the various variables and update only one at a time given the current solutions of the remaining variables. In our case, this splits the optimization into three parts: (i) finding the most likely latent state configuration given the intermediate solutions of the latent-class SVDD and log-linear model part, (ii) finding the optimal solution for the latent-class SVDD part given the current latent state configuration, and (iii) finding the optimal parameter for the log-linear model given the current latent state configuration. Luckily, sub-problems (ii) and (iii) are convex problems and any local solution will also be a global optimal solution.

If $\nu=1$, the optimal solution can be found analytically by $\bc_k = 1/n_k \sum_i \mathbf{1}[h_i=k]\phi(\bx_i)$.
\begin{algorithm}[t]
    \begin{algorithmic}[t]
        \STATE input data $X$, number of components $K$, mixture parameter $0 \leq \theta \leq 1$, and fraction of anomalies $0<\nu \leq 1$
        \STATE set $t=0$ and initialize $\bc_k^t$ and $\bv^t$ (e.g., randomly)
        \REPEAT
            \STATE t:=t+1
            \STATE Infer latent states $H^t$ using intermediate solutions $\bv^{t-1}$ and $\bc_k^{t-1}$
			\STATE Update $\bc_k^{t}$ given $H^t$ %$\bc_k^{t} = 1/n_k \sum_i \mathbf{1}[z^t_i=k]\phi(\bx_i)$
			\STATE Update $\bv^{t}$ given $H^t$
        \UNTIL $H^t =  H^{t-1}$
        \STATE Calculate final anomaly scores $s(\bx_i) = \|\bc_{h^t_i} - \phi(\bx_i) \|^2$ 
    \end{algorithmic}
    \caption{Latent-class contextual anomaly detector (LCCAD)\label{alg:lccad}}
\end{algorithm}

For sub-problem (i) and tree-like structures, optimal solutions can be found using e.g. belief propagation algorithms or linear program approximations (cf. \cite{Wainwright2007} for a comprehensive discussion). For arbitrary structures, loopy belief propagation approximation \cite{Weiss2000}, where each $\hat{H}$ is sequentially updated given the states of its neighbors, is a powerful and fast method to solve the problem.
This algorithm works by iteratively sending messages $M_{ij}(s)$ from node $i$ to node $j$ (in state $s$) in the proximity of its location:
\begin{align*}
% messages
M_{ij}(s) &\leftarrow \varepsilon + \max_{t} \iota_{ij}(s,t) + \vartheta_i(t) + \sum_{k \in N(i) / j} M_{ki}(t) \;,
\end{align*}
where $\varepsilon$ is some normalization constant, $N(i)$ denotes the set of neighboring nodes of node $i$ and 
\begin{align*}
\iota_{ij}(s, t) &= (1-\theta) \bv_{st},  \\
\vartheta_i(t) &= (1-\theta) \langle \bv_t, \phi(\x_i)\rangle 
- \frac{\theta}{n_t \nu} \ell(\|\bc_t - \phi(\bx_i) \|^2 - t_t)\;.
\end{align*}
After convergence, {max-marginals $\mu_i(s)$ can be computed as follows:}
\begin{align*}
% max-marginals
\mu_{i}(s) &\leftarrow \varepsilon + \max_{t} \vartheta_i(t)+\sum_{k \in N(i)} M_{ki}(t) \; .
\end{align*}
Finally, backtracking using the max-marginals reveals the latent states per node.
We empirically found that LBP approximations are fast, converging within a few iterations, and give reasonable results. The final optimization problem is given in Algorithm~\ref{alg:lccad}. For relations to other anomaly detection algorithms and special cases see Supplement. 

%--------------------------------------
\section{Explaining Latent-class Contextual Anomalies}\label{sec:explanation}
%--------------------------------------

In addition to a reliable detection of anomalies it is important to explain and understand why a point has been detected as anomalous \cite{DBLP:conf/icdm/MicenkovaNDA13,
DBLP:journals/corr/abs-1711-10589,
DBLP:journals/corr/KauffmannMG18}. The explanation method we present here identifies to what extent each input feature is relevant to the decision. Denoting by $o(x_1,\dots,x_d)$ the predicted outlierness of a data point $\x \in \mathbb{R}^d$, we explain by producing a decomposition $o(x_1,\dots,x_d) = R_1 + \dots + R_d$ where $R_i$ is the relevance of variable $x_i$.

A number of techniques have been proposed for decomposing machine learning predictions in terms of input variables \cite{DBLP:conf/aaai/PoulinESLGWFPMA06, DBLP:conf/cidm/LandeckerTBMKB13, Bach2015, DBLP:journals/pr/MontavonLBSM17, DBLP:conf/icml/ShrikumarGK17}. 
In the SVDD model of Section \ref{sec:method}, outlier scores to be explained are squared-norms with an offset and a rectification function. While the squared norm has a simple additive structure, it applies to the feature space,  the dimensions of which,  are not interpretable. Instead, we would like a decomposition in terms of the input variables. As a first step, we note that SVDD applied to the feature space of a Gaussian kernel is strictly equivalent to the One-Class SVM with the same kernel \cite{estimating-the-support-of-a-high-dimensional-distribution}. Kauffmann et al.\ \cite{DBLP:journals/corr/KauffmannMG18} proposed a method called one-class deep Taylor decomposition (OC-DTD) which explains One-Class SVM predictions in terms of input variables. The method is based on a deep Taylor decomposition (DTD) explanation framework that was first developed in the context of deep neural network classifiers \cite{DBLP:journals/pr/MontavonLBSM17}. We describe the main idea of the OC-DTD in the following. Let
\begin{align}
f_k(\bx) = \sum_{\{j\mid h_j=k\}}\alpha_{j} \mathbb{K}(\bx,\bx_{j}),\quad k=1,\ldots,K
\end{align}
be the $K$ SVM discriminants, where $\mathbb{K}(\bx,\cdot)$ is the kernel function associated to the feature map $\phi(\bx)$.
When $\mathbb{K}$ is Gaussian, we can measure the degree of outlierness by the monotonically decreasing function $o_k(\bx) = -\log(f_k(\bx))$. OC-DTD first writes this function as a two-layer neural network: Layer $1$ is a mapping to the effective distances $d_{j}(\x) = \|\Delta_{j}\|^2\slash 2 - \log \alpha_j$, where $\Delta_{j} = (\bx-\bx_j)/\sigma$ is the scaled difference from the support vector, and where non-support vectors ($\alpha_j=0$) are effectively infinitely far. Layer $2$ applies a soft min-pooling to these effective distances: $o_k((d_j)_j) = -\log \sum_j \exp(-d_{j}) \approx \min_j d_{j}$. That is, a point is an outlier if no support vector is nearby. Application of the deep Taylor decomposition to this two-layer neural network yields the relevance scores:
\begin{align}
R_i =
\sum_{\{j\mid h_j=k\}}\frac{[\Delta_{j}]^2_i}{\|\Delta_{j}\|^2}
\cdot
\frac{\alpha_{j} \mathbb{K}(\bx,\bx_{j})}{\sum_{j'} \alpha_{j'} \mathbb{K}(\bx,\bx_{j'})}
\cdot
\min\left(o_k(\bx),\|\Delta_{j}\|^2\right)
\label{eq:relevance}
\end{align}
While we refer the reader to \cite{DBLP:journals/corr/KauffmannMG18} for a derivation of $R_i$, inspecting the product structure in Eqn \eqref{eq:relevance} gives the following interpretation: An input feature $x_i$ is relevant to anomaly if (1) it differs to the support vector more than other features, (2) the support vector is among the nearest support vectors, and (3) the point is an outlier within its assigned cluster $k$.
%--------------------------------------------
\section{Empirical Evaluation}\label{sec:experiments}
%--------------------------------------------
We divide the empirical evaluation into three distinct parts. First, we verify the usefulness of our method in the described latent-class contextual anomaly setting where we have full control over data generation and hence, the latent variables as well as the corresponding anomalies. We compare the anomaly detection accuracy of our method against several state-of-the-art anomaly detection methods. Furthermore, we test the ability to uncover the underlying cluster structure and compare the results against $k$-means clustering.
In the second part, we apply our method to the well-known synthetic reservoir dataset. Here, we can only quantify the ability to identify the underlying clustering structure as no ground truth anomalies are known. 
Finally, we unleash the full potential of our methodology by applying it to a large-scale real-world reservoir data set where no ground truth, either in form of anomaly labels nor in form of latent variables, is given. We let a domain expert assess the solution and explanation produced by our method.

Throughout the empirical evaluation, we restrict the model to $\nu=1$. Furthermore, we measure anomaly detection accuracy in area under the ROC curve (AUROC) and clustering accuracy in adjusted Rand index (ARI). We empirically found that automatically adjusting the regularization hyperparameter $\gamma$ such that $\|\bv^1\| = 1$ gives reasonable results across multiple applications.
The use of Gaussian kernels (cf.\ \cite{MueMikRaeTsuSch01} for an introduction to kernel methods) has been observed to achieve good results for one-class classification \cite{VertVert06}. In order to use our model as defined in Def.\ \ref{def:lccad} we employ a feature transformation that resembles Gaussian kernels \cite{RahimiRecht2009}. We can rewrite the kernel machine as a neural network \cite{DBLP:journals/corr/KauffmannMG18} and use our explanation method from Section~\ref{sec:explanation}.

\begin{figure}
\includegraphics[width=\textwidth]{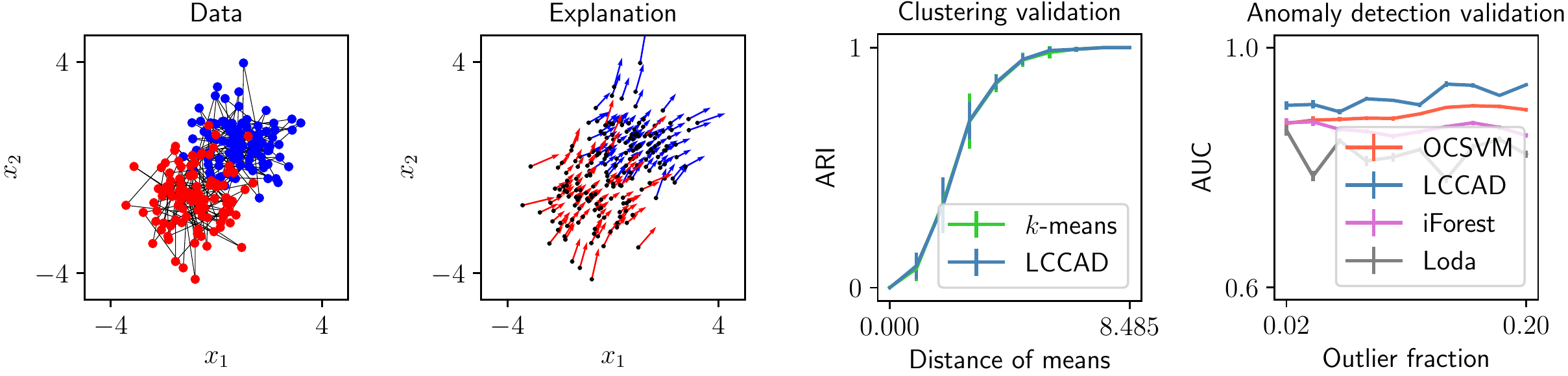}
\caption{From left to right: Examplary toy data set; OC-DTD explanation of latent class anomalies, ARI for LCCAD and $k$-means for increasing mean-distances on the toy-data with standard deviation on ten randomly generated test sets; AUC for several contamination rates along with standard deviation on ten (randomly generated) test sets.}\label{fig:corr_toy}
\end{figure}

%--------------------------------------------
\paragraph{Verification on Artificial Data}
%--------------------------------------------
We test the ability to find anomalies and corresponding clustering structure in latent-class contextual settings on an artificially generated two-dimensonal dataset shown in \Cref{fig:corr_toy} (left). The two classes are generated by
\begin{align*}
\x_i \sim \boldsymbol{\mu}_c + \mathcal{N}(\boldsymbol{0}, \sigma_c^2I)
\end{align*}
with classes $c\in\{1,2\}$ respectively. Therefore, the ground truth anomaly score will be defined as the amount of distortion that is due to the Gaussian noise,
\begin{align*}
o_c(\x_i) = \frac{\|\x-\boldsymbol{\mu}_c\|^2}{2\sigma_c^2}.
\end{align*}
On this dataset, we expect a baseline anomaly model, where the latent classes (depicted by colors red and blue) are not used, to identify as outlier only points outside the mixed distribution. On the other hand, the LCCAD approach should find additional outliers located within the other class.

As a preliminary step, in order to verify the clusters built by LCCAD, we perform a comparison with the standard $k$-means clustering algorithm. The third plot of \Cref{fig:corr_toy} shows the adjusted rand index for various distances $\|\boldsymbol\mu_1-\boldsymbol\mu_2\|$ between the means of the two latent classes. On the left, classes overlap completely and no separation is possible. On the right side, classes are completely separated. We observe that LCCAD performs comparable to $k$-means in this task. For a fair comparison, we performed a nested cross-validation and show the performance on ten unseen test sets per distance for both methods.

Now that the clustering performed by LCCAD has been validated, we assess the ability of our method to detect anomalies reliably. The last plot of \Cref{fig:corr_toy} shows the AUC for several anomaly thresholds on $o_c(\x_i)$. Also here, results are shown for ten unseen test sets per outlier fraction. Our proposed LCCAD method is clearly superior to all baselines in this task.

%--------------------------------------------
\paragraph{Evaluation on Synthetic Reservoir Data}
%--------------------------------------------
\begin{wrapfigure}{r}{0.4\textwidth}
\vspace{-.5cm}
  \begin{center}
  \includegraphics[width=0.3\textwidth]{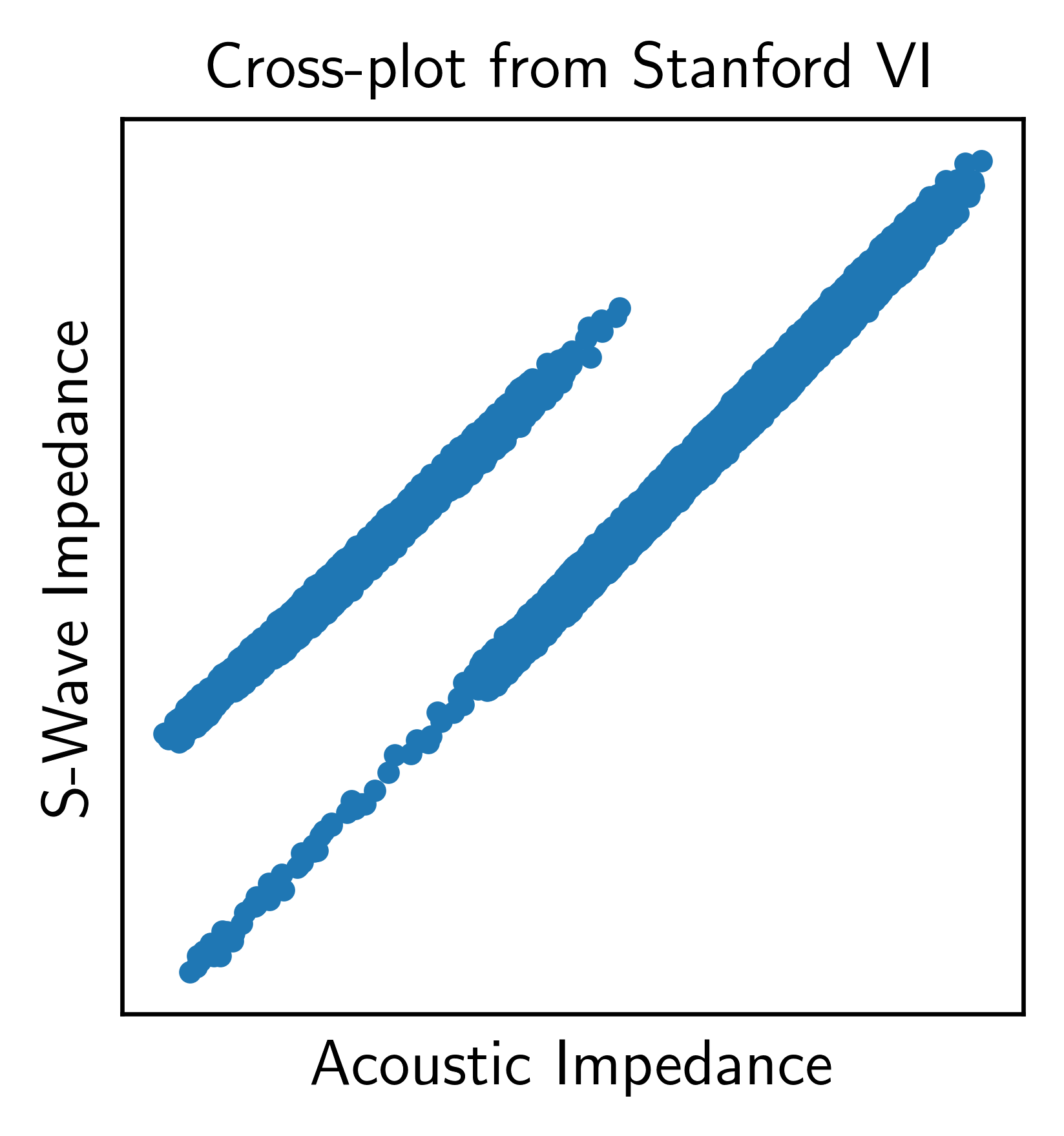}
\label{fig:cp-synth}
  \end{center}
\vspace{-.2cm}
 \caption{Cross-plot of acoustic impedance $AI$ versus S-wave impedance $SI$ from synthetic seismic data.}
 %\vspace{-.2cm}
\end{wrapfigure}
In this section, we measure the ability to find grouped structures on synthetic reservoir data. We further use explanation methods to verify anomaly findings due to lack of ground truth data.
We use the synthetic 3D reservoir benchmark data set Stanford VI \cite{Castro2005} ($150 \times 200 \times 200$ voxels),
which was created through realistic geological modeling.
It contains two facies: the sand channels (blue in Fig.~\ref{fig:layer_explained} upper, left) and the background shale (red). Due to the vertical low resolution during the seismic acquisition process \cite{Deutsch2002}, we simplify our setting by only considering connections in the horizontal slices. So, from each of those volumes, we extract $150 \times200$ horizontal slices,
and assume that the whole impedance data is available
as the input (center and right images in the top row of Fig.~\ref{fig:layer_explained}). Since facies are available as a ground truth clustering for the data set, we calculated ARI for $k$-means and LCCAD and found an index of 0.88 for both methods without significant differences throughout the data set. This is no surprise if we look at the cross-plot for both available features, \Cref{fig:cp-synth}. Our goal is to find extremal contextual (spatial connections) anomalies by inferring the latent class structures (facies).

\begin{wrapfigure}{l}{0.4\textwidth}
\vspace{-.0cm}
  \begin{center}
  \includegraphics[width=0.3\textwidth]{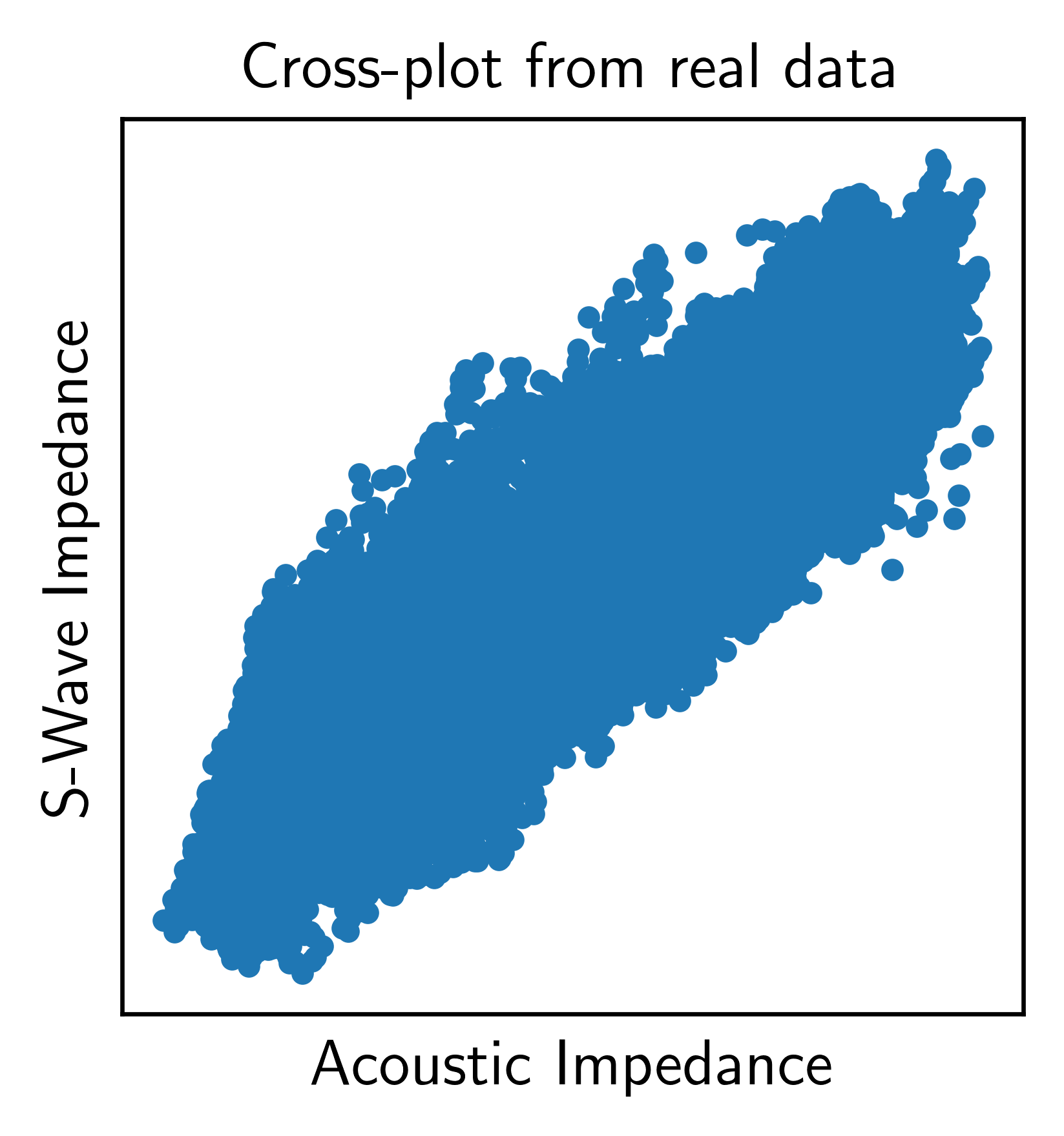}
\label{fig:cp-real}
  \end{center}
  \vspace{-.2cm}
 \caption{Cross-plot of acoustic impedance $AI$ versus S-wave impedance $SI$ from the real seismic data.}
 \vspace{-1cm}
\end{wrapfigure}

We compare our findings (center row in Fig.~\ref{fig:layer_explained}) 
against baseline competitors LODA, vanilla OC-SVM, and isolation forests (bottom row in Fig.~\ref{fig:layer_explained}) which do find similar global anomalies (in different scales) within the shale facies as well as the sand channels. In contrast, the results of our method suggest, after almost perfect reconstruction of the latent-class configuration, that there is only a single spot of anomalies within the shale facies that significantly deviates from the remainder of the data. 
Moreover, the decomposition into input feature slices (center row center and right) according to our proposed explanation method, suggests that seismic impedance $R_{SI}$
is the origin of the anomalous signal. If we assume that the discovered latent states correspond to the ground 
truth, we find that some features are better suited for 
inference of latent states, since they contain only fewer 
anomalies. Interestingly, the found anomalous spot has 
meanwhile been assessed to be an echo from a deeper 
sand structure by a geophysics expert. 
%--------------------------------------------
\paragraph{Application to Real-world Reservoir Data}
%--------------------------------------------
We apply our method to a real petroleum reservoir, located in the offshore coast of Brazil. It covers an area of approximately 100 square kilometers, with 460 meters in depth. The data in this region comprises a 3D volume with $313 \times 549 \times 74$ voxels containing acoustic and S-Wave impedance samples. \Cref{fig:cp-real} shows a cross plot the particular slice of our analysis.
This data contains \emph{truly} labeled data from only four wells, with which no \emph{general-purpose} machine learning method can cope. Hence, any finding need careful \emph{manual} assessment by some domain expert.
For our analysis, we assume a three latent states as relevant for detection of contextual anomalies. This corresponds also to the number of facies in this data set \cite{DBLP:journals/gandc/LimaGVVMN17}. Similar to the results obtained on the simulated data, we can use the feature-wise explanation for analysis of the origin of detected anomalies.
Firstly, we see that latent states capture ridge spatial structures which partially overlap with findings from related semi-supervised methods (cf.\ \cite{DBLP:journals/gandc/LimaGVVMN17}).
Secondly, one can find from the explanations $R_{AI}$ and $R_{SI}$, that anomalous fragments are mostly due to distortion in S-wave impedance features. This is similar to our findings on the synthetic data set. 

\begin{figure}
\centering
\begin{minipage}{.48\textwidth}
  \centering
\includegraphics[width=\textwidth]{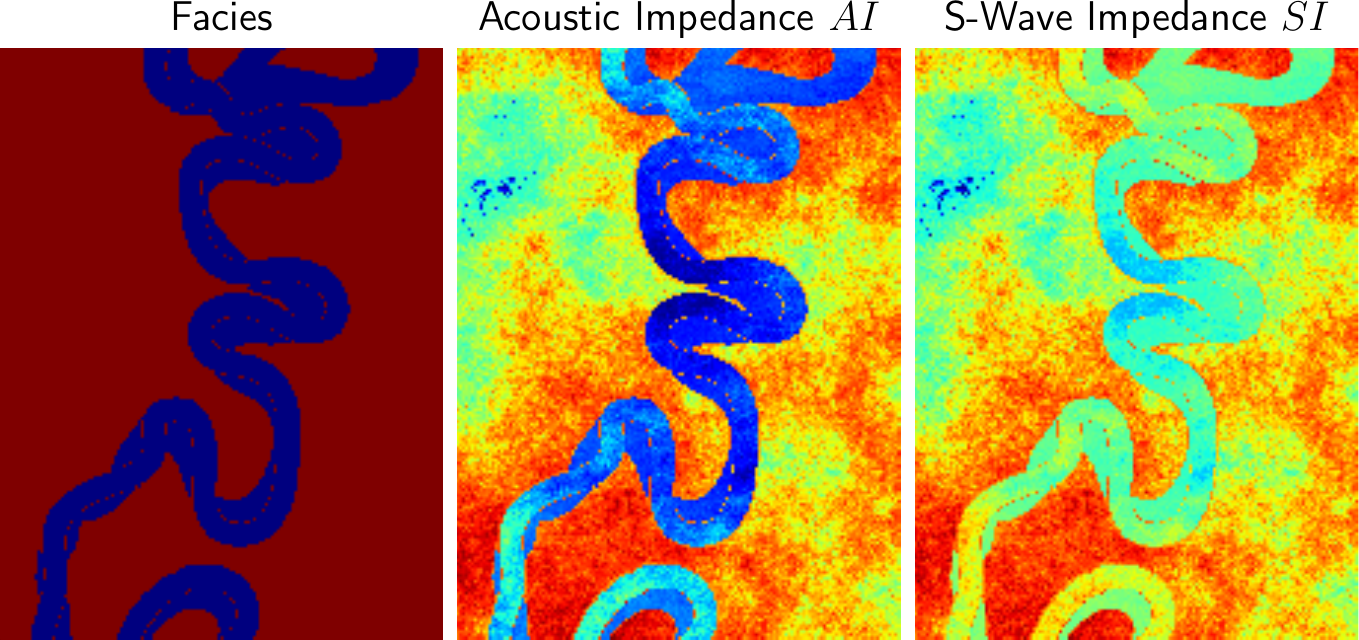}\\
\includegraphics[width=\textwidth]{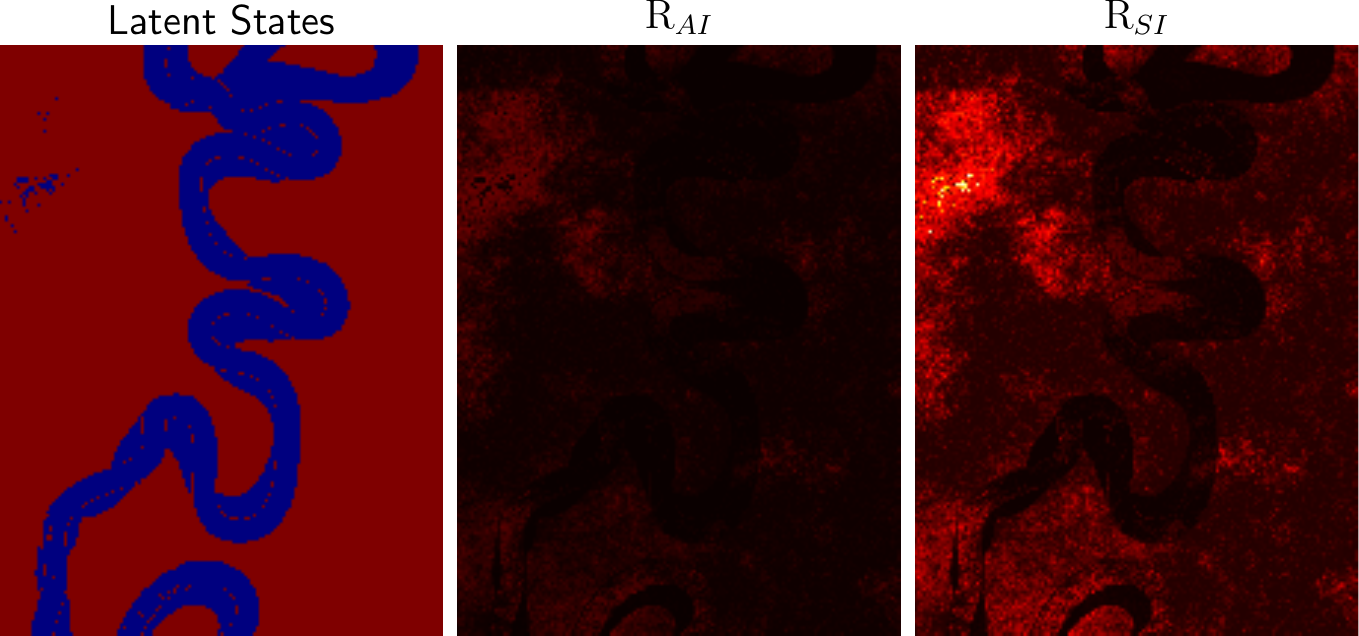}\\
\includegraphics[width=\textwidth]{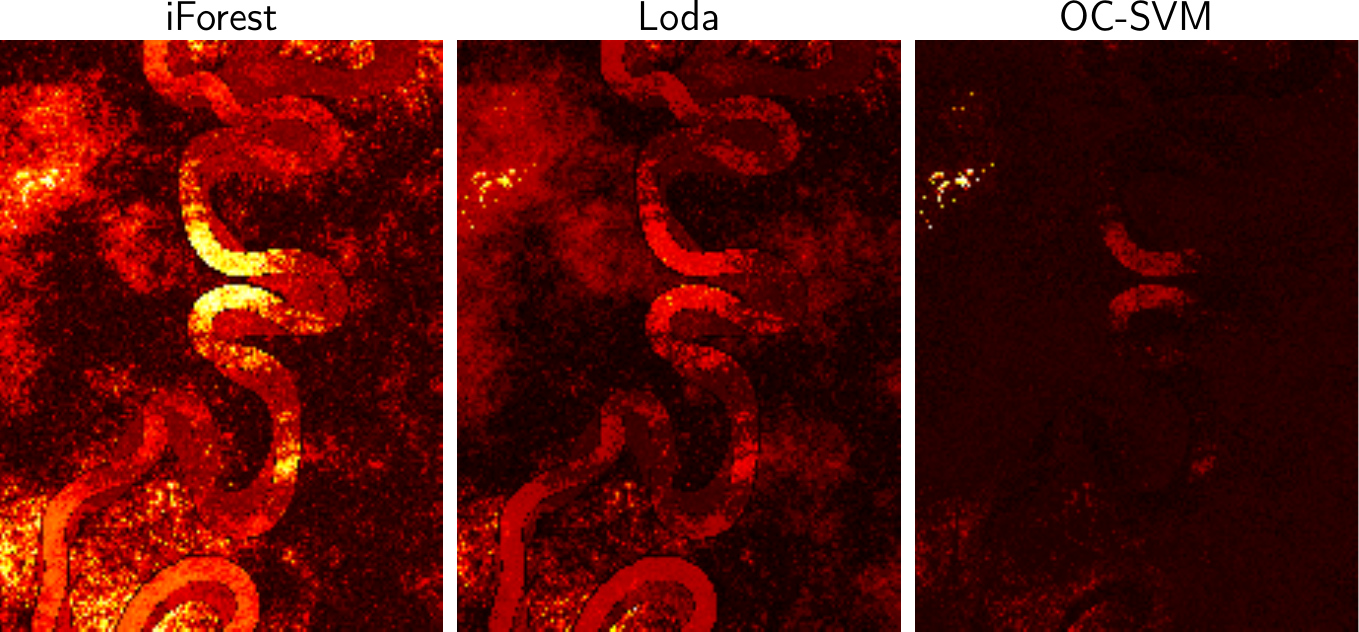}
  \caption{Synthetic data set: {\emph Top:} Facies and features $AI$ and $SI$; {\emph middle:} Latent states by LCCAD and feature-wise anomaly explanation by OC-DTD; {\emph bottom: baseline anomaly detection methods}}
  \label{fig:layer_explained}
\end{minipage}%
\rulesep
\begin{minipage}{.48\textwidth}
  \centering
\includegraphics[width=\textwidth]{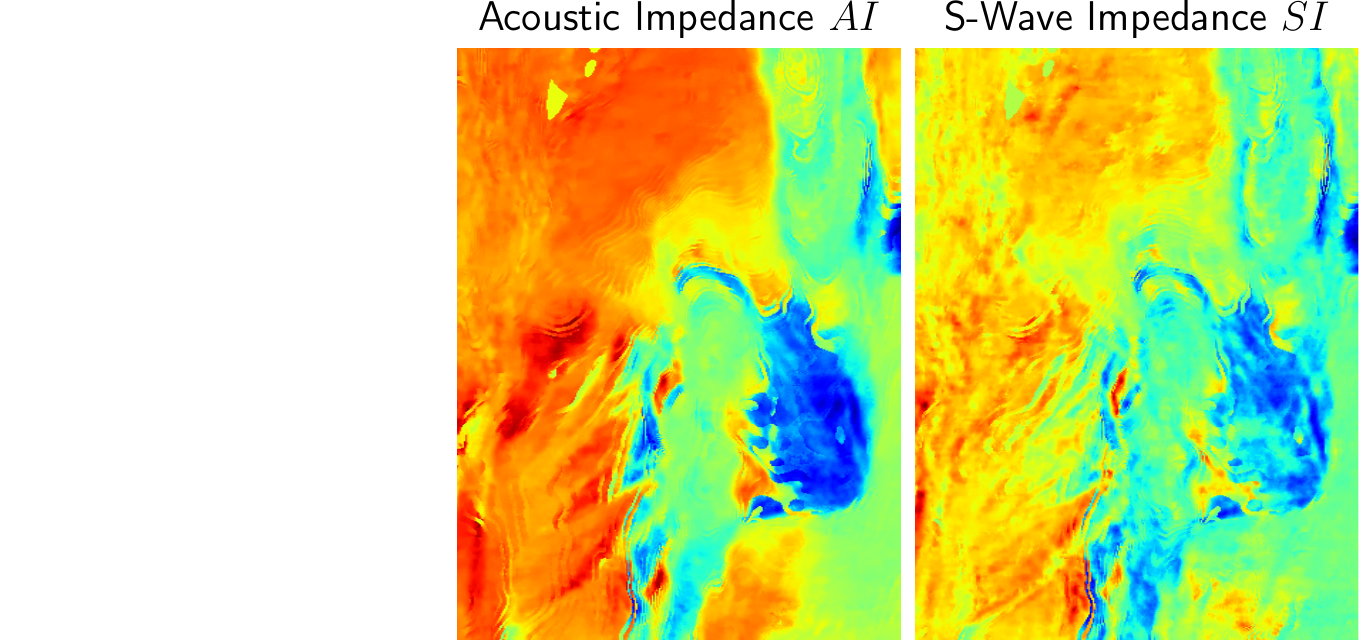}\\
\includegraphics[width=\textwidth]{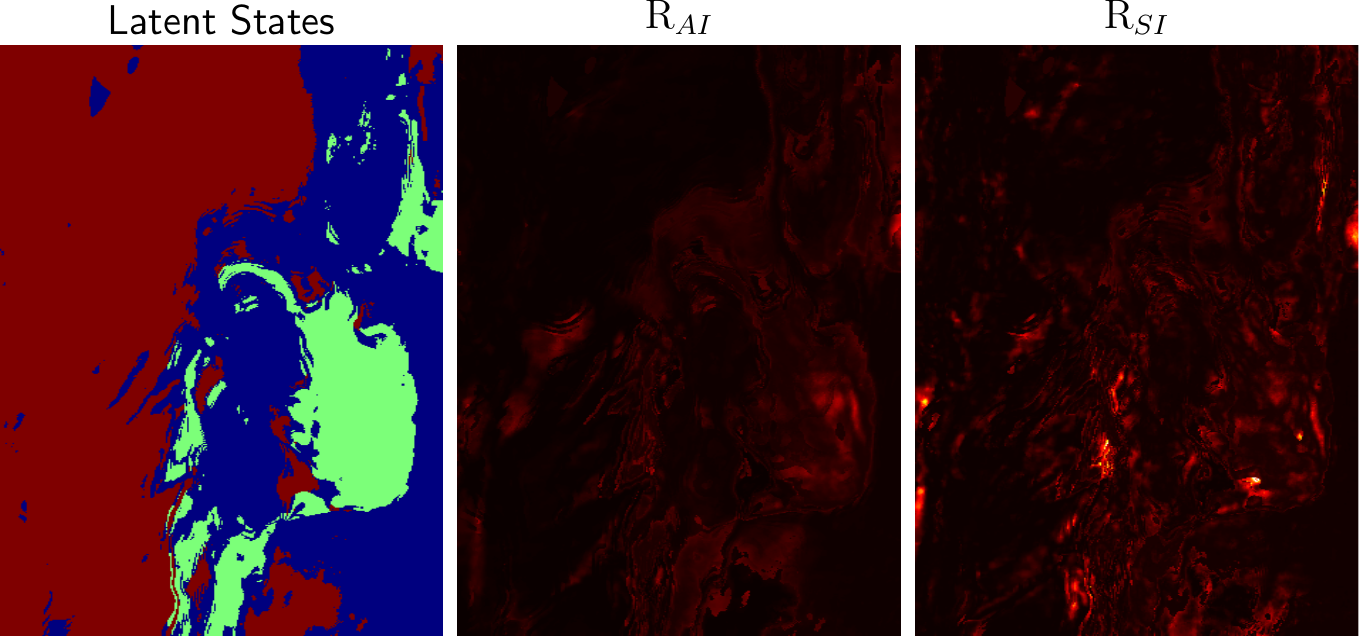}\\
\includegraphics[width=\textwidth]{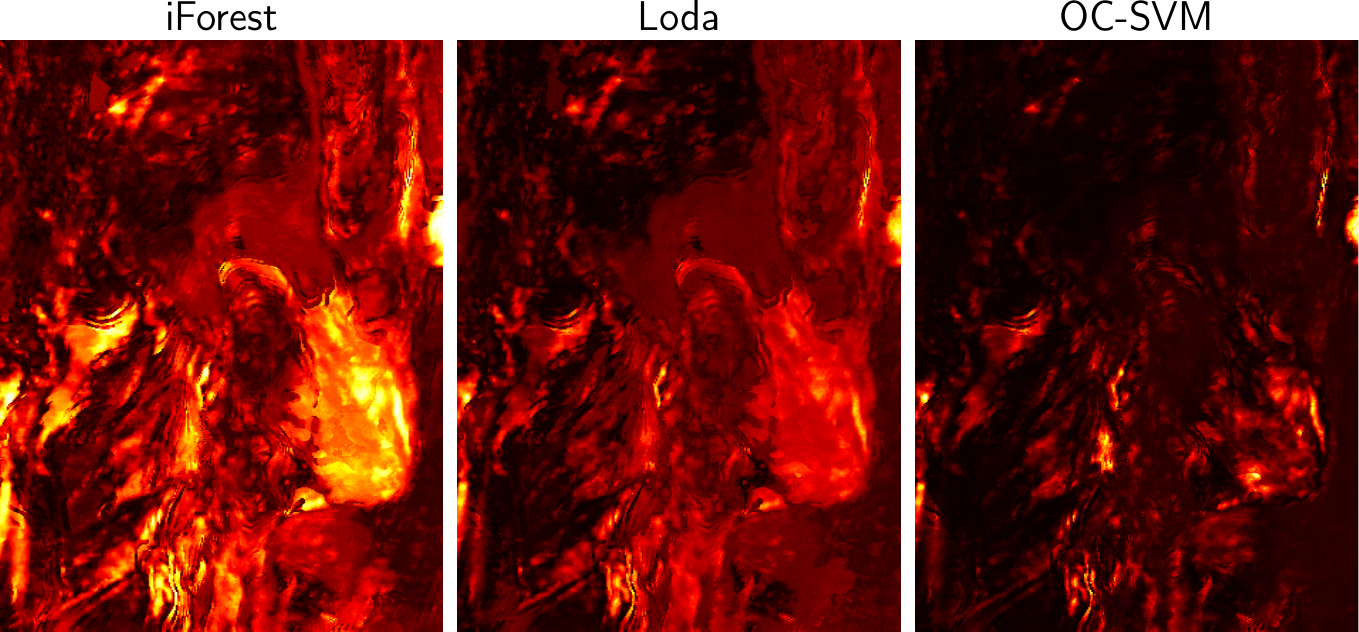}
  \caption{Real seismic data set: {\emph Top:} Features $AI$ and $SI$; {\emph middle:} Latent states by LCCAD and feature-wise anomaly explanation by OC-DTD; {\emph bottom: baseline anomaly detection methods}}
  \label{fig:layer_explained}
\end{minipage}%
\end{figure}

%--------------------------------------------
\section{Conclusion}\label{sec:conclusion}
%--------------------------------------------

Analyzing real world data requires rich models that can learn and model the underlying complexity and structure. Outliers can significantly disrupt this modeling process and they are particularly harmful when few labels exist and the data has a rich structure including latent states. We have contributed to address these challenges by proposing a novel outlier detection model for structured data with intrinsic unknown latent states. Toy data show the usefulness and unique capabilities of our novel model; clearly providing a scenario where only our model can be successful as effectively no competitors exist. To demonstrate that such scenarios exist and are relevant in the real world, we study geophysics data from oil exploration. Here we can show that structured outliers with a latent state can help to accurately detect the unknown facies structure. An important additional finding is that we can adapt explanation techniques originally defined for supervised learning to our unsupervised structured outlier detection case. This allows not only detecting but also to explain and to visualize why outliers are considered outliers by our model -- an immense progress for a geophysics practitioner.

Future work will explore this framework in other scientific applications beyond geophysics. In particular in the medical domain outliers in complex data with latent states are of high interest, they may e.g. be the responders to drugs or e.g. long-term survivors.

%--------------------------------------------
\subsubsection*{Acknowledgments}

This work was supported by the German Research Foundation (GRK
1589/1) by the Federal Ministry of Education and Research (BMBF) under the BMBF project ALICE II, Autonomous Learning in Complex Environments (01IB15001B),
and the project Berlin
Big Data Center (FKZ 01IS14013A).

%--------------------------------------------

%--------------------------------------------
%\subsubsection*{References}
%--------------------------------------------
%\newpage
\bibliographystyle{abbrv}
{\small
\bibliography{nico,la,explanation}
}
%
%--------------------------------------------
\clearpage
\setcounter{page}{1}
\appendix

\section{Supplement}

%--------------------------------------
\subsection{Discussion, Relations, and Special Cases}
%--------------------------------------
Our LCCAD method performs unsupervised anomaly detection. It is inspired by the supervised algorithm of transductive conditional random field regression (TCRFR)~\cite{GoeLimVarMueNak17};  LCCAD being  easier to use as it has  less free parameters. Moreover, the latent-class SVDD part of LCCAD is inspired by ClusterSVDD as given in \cite{GoeLimMueKloNak2017}.

Notable special cases of our LCCAD approach include vanilla SVDD~\cite{TaxDui04} which is recovered when $K=1$, $\theta=1$, and $\ell(x)=\max(0, x)$. Conditional random fields~\cite{LafMccPer01} are recovered when $K > 1$ and $\theta=0$ (however, without any provided latent states for parameter estimation). Moreover, if 
$K > 1$, $\theta=1$, $\ell(x)=\max(0, x)$, and $\nu=1$, our proposed method LCCAD becomes equivalent to $k$-means~\cite{MacQueen1967}.

LCCAD assumes that (i) useful dependency structure is given an (ii) that latent-class dependencies exists (cf. Fig.~\ref{fig:lccad-model}). Especially if (i) is not fulfilled, our method could easily get less accurate than its base model SVDD since it has no structure information to capitalize from. Further, it is worth mentioning that contexts of out-of-sample data points can not be readily inferred without re-training. Moreover, due to the increased complexity and non-convexity, runtime performance is much slower for LCCAD than for vanilla SVDD.  
\end{document}